\documentclass[conference]{IEEEtran}
\IEEEoverridecommandlockouts
\usepackage{cite}
\usepackage{amsmath,amssymb,amsfonts}

\usepackage{graphicx}
\usepackage{textcomp}
\usepackage{xcolor}
\def\BibTeX{{\rm B\kern-.05em{\sc i\kern-.025em b}\kern-.08em
    T\kern-.1667em\lower.7ex\hbox{E}\kern-.125emX}}

\usepackage{algorithm}
\usepackage[noend]{algpseudocode}

\begin{document}

\title{CIMRL: Combining IMitation and Reinforcement
Learning for Safe Autonomous Driving\\
}


\author{\IEEEauthorblockN{1\textsuperscript{st} Jonathan Booher\textsuperscript{*}}
\and
\IEEEauthorblockN{2\textsuperscript{nd} Khashayar Rohanimanesh\textsuperscript{*}}
\and
\IEEEauthorblockN{3\textsuperscript{rd} Junhong Xu\textsuperscript{*}}
\and
\IEEEauthorblockN{4\textsuperscript{th} Vladislav Isenbaev\textsuperscript{**}}
\and
\IEEEauthorblockN{5\textsuperscript{th} Ashwin Balakrishna\textsuperscript{**}}
\and
\IEEEauthorblockN{6\textsuperscript{th}  Ishan Gupta\textsuperscript{**}}
\and
\IEEEauthorblockN{7\textsuperscript{th} Wei Liu\textsuperscript{**}}
\and
\IEEEauthorblockN{8\textsuperscript{th}  Aleksandr Petiushko\textsuperscript{**}}
}


\maketitle

\begin{center}
    \textbf{Nuro, Inc.}
\end{center}

\begin{abstract}
Modern approaches to autonomous driving rely heavily on learned components trained with large amounts of human driving data via imitation learning. However, these methods require large amounts of expensive data collection and even then face challenges with safely handling long-tail scenarios and compounding errors over time. At the same time, pure Reinforcement Learning (RL) methods can fail to learn performant policies in sparse, constrained, and challenging-to-define reward settings such as autonomous driving. Both of these challenges make deploying purely cloned or pure RL policies in safety critical applications such as autonomous vehicles challenging. In this paper we propose {\em Combining IMitation and Reinforcement Learning} (CIMRL) approach --- a safe reinforcement learning framework that enables training driving policies in simulation through leveraging imitative motion priors and safety constraints. CIMRL does not require extensive reward specification and improves on the closed loop behavior of pure cloning methods. By combining RL and imitation, we demonstrate that our method achieves state-of-the-art results in closed loop simulation and real world driving benchmarks.

\end{abstract}

\newcommand\blfootnote[1]{%
  \begingroup
  \renewcommand\thefootnote{}\footnote{#1}%
  \addtocounter{footnote}{-1}%
  \endgroup
}
\newcommand{\Q}{\mathcal{Q}}

\blfootnote{* \{jbooher, krohanimanesh, juxu\}@nuro.ai}
\blfootnote{** All work related to this paper performed while employed by Nuro, Inc.}



\section{Introduction}
\label{sec:introduction}
\begin{figure*}[h]
\centering
    \includegraphics[width=14cm]{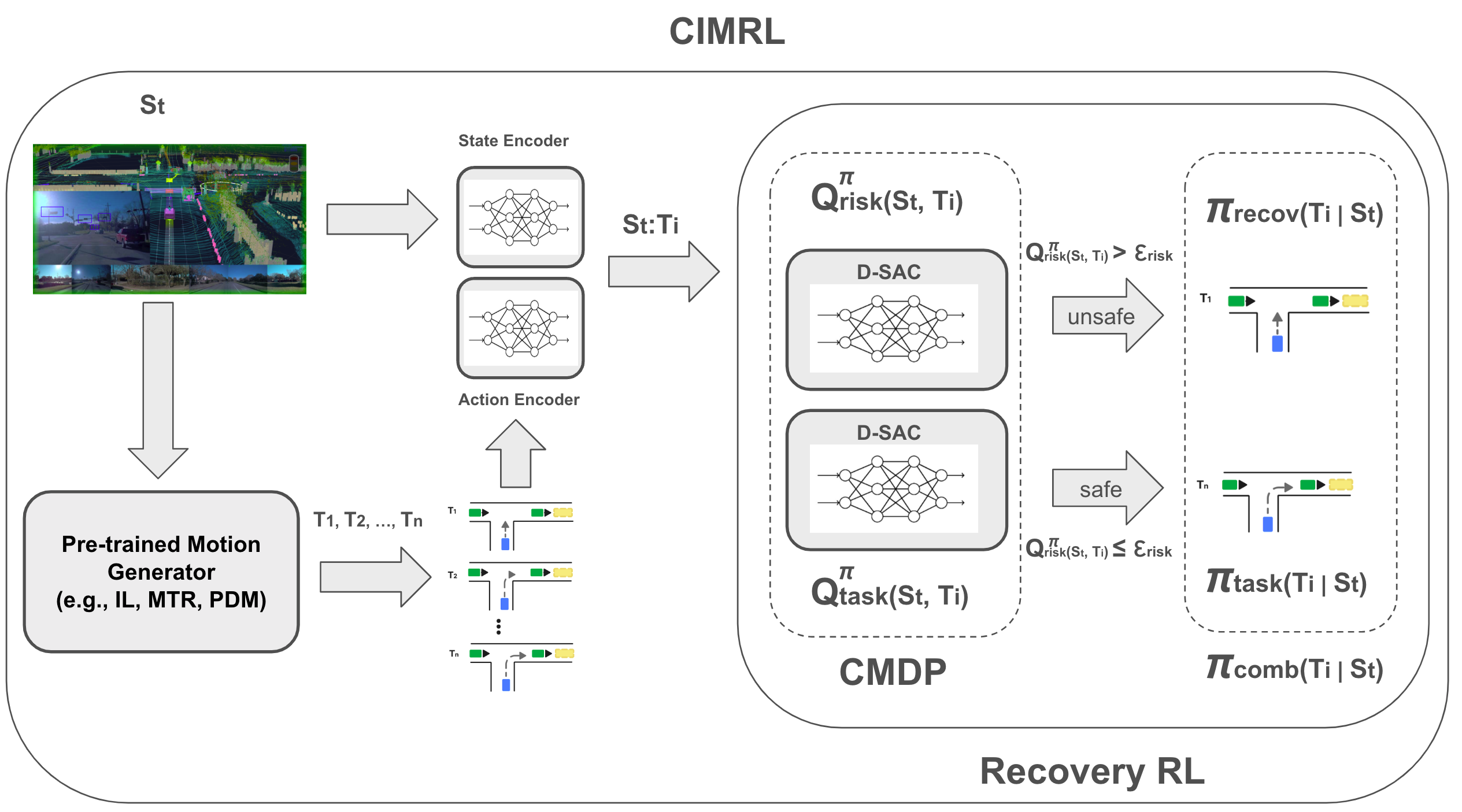}
    \caption{Illustration of the CIMRL algorithm. The model combines imitation learning with safe reinforcement learning by restricting the action space to an efficient support derived from the motion prior generated by a pretrained imitation learning model. State and action are encoded via deep neural networks, concatenated, and used to predict both task and risk values. The model is initially trained in simulation and then deployed in real-world environments, ensuring robust and scalable performance. We use the $Q_{risk}$ estimation to identify safe actions based on a risk threshold. If such actions exist, we use the task policy $\pi_{task}$ to select exclusively from the safe actions. However, should there be no safe actions available, we fall back to using the recovery policy $\pi_{recov}$ which is optimized to guide the agent back to a safe state.}
    \label{fig:cimrl-arch}
\end{figure*}
The development of self-driving cars has been mainly driven by advancements in three key areas: perception, prediction, and planning. These components form the foundation of modern the self-driving car stack (SDC), enabling autonomous vehicles to navigate safely and efficiently. In recent years, deep learning has significantly transformed the design of SDC, particularly revolutionizing the first two components: perception \cite{li2022bevformer,DBLP:journals/corr/abs-1711-08488,DBLP:journals/corr/abs-1711-06396}, and prediction \cite{DBLP:journals/corr/abs-1910-05449,nayakanti2022wayformer,DBLP:journals/corr/abs-2008-08294}. Most of these prior works leverage supervised learning on large, manually or auto-labeled datasets and curated expert driving behavior. Methods such as behavior cloning (BC)\cite{pomerleau:alvinn,bojarski2016endendlearningselfdriving} and other imitation learning (IL)\cite{ross2011reductionimitationlearningstructured,codevilla2018endtoenddrivingconditionalimitation} methods have shown promise at effectively controlling ego behavior, in particular replicating human-like driving behavior. However, these methods can struggle in scarce data regimes like those encountered in safety-critical situations, and are known to suffer from compounding errors in out of distribution scenarios \cite{ross2011reductionimitationlearningstructured}. 
On the other hand, pure reinforcement learning (RL) approaches to autonomous driving~\cite{Learning-to-Drive-in-a-Day-ICRA2019, lu2023imitation} while leading to increased robustness, and exploration, face challenges due to the high-dimensional state and action spaces, the need for manually crafted multi-criteria reward functions, the lack of safety-critical guarantees (due to multiple competing criteria in the reward function), and the tendency to learning policies that may not naturally mirror human driving behavior \cite{lu2023imitation}. Therefore, the most effective learning method for ego planning remains an unsolved problem. 



In this work, we aim to combine the strengths of both approaches by addressing the problem through four key criteria: (1) safety constraints and guarantees, ensuring that critical safety standards (e.g., prioritizing human safety above all) are not compromised in achieving task-level objectives, even in rare, catastrophic long-tail events; (2) robustness, particularly in long tail and challenging scenarios in the absence of abundant of expert data; (3) generation of natural driving behavior that closely mirrors human driving; and (4) scalable closed-loop learning in simulation and transfer to real world, capable of coping with large, complex multi-dimensional action spaces and the intricate nature of agent-environment interactions.
Figure\ref{fig:cimrl-arch} presents an overview of or approach, where we use a Safe-RL framework adapted from the {\em Recovery RL} framework \cite{DBLP:journals/corr/abs-2010-15920} to address the first two criteria. More specifically, to model explicit risk severity, we train two distinct policies — one dedicated to optimizing task reward, and the other focused on minimizing risk severity. Acknowledging the continuous nature of driving and the scarcity of risk events, we employ the Tree Backup \cite{Asis2017MultistepRL} algorithm to aid in effectively estimating the task and risk value functions (e.g., $Q_{task}$ and $Q_{risk}$). The final control policy (see Figure~\ref{fig:cimrl-arch}) can be viewed as a two level hierarchical RL \cite{sutton1999between} where the top level policy implements a a strategy that dynamically switches control between the lower-level task and risk policies: when safety constraints are at risk of being violated (as measured by $Q_{risk}$), control is handed over to the risk minimization policy, referred to as the recovery policy in this framework, which focuses on minimizing risks during this phase. Once the recovery policy guides the agent to the safety zone, the high level policy switches the control to the task optimization policy which learns to maximize the task reward during this phase. Framing the problem as safe RL allows us to develop safe and robust policies, particularly for long-tail and challenging scenarios. These scenarios can often be generated and tested more cost-effectively in simulation without the risk of catastrophic outcomes, allowing the learning agent to perform safe exploration, and ensuring greater reliability before deployment in real-world environments.

When using online RL methods, it is important to carefully design the state and action spaces, and rewards to elicit the desired behavior from the resulting policy. In domains like self-driving, modeling the complex transition dynamics of the environment and designing intricate reward structures are often viewed as challenges to online RL. One naive method to train RL-based policies is to model the action space as only the immediate action for the next invocation \cite{lu2023imitation}. While appealing in its simplicity, we demonstrate that this choice is not optimal. To address this, we propose to leverage state-of-the-art trajectory prediction models \cite{DBLP:journals/corr/abs-1910-05449,varadarajan2021multipathefficientinformationfusion,shi2022motion} as imitative motion priors and learn a discrete policy to select amongst these quality candidates (see Figure\ref{fig:cimrl-arch}). This approach significantly reduces the learning complexity by replacing the complex trajectory generation problem during closed loop training by a simpler trajectory selection problem. By leveraging this discretization method, we can narrow the exploration space, ensure human-like policies, and incorporate domain knowledge to generate feasible, traffic-rule compliant backup plans, hence addressing the third and fourth criteria mentioned above. We also demonstrate that even with simplified transition dynamics and reward design, effective results can still be achieved and successfully transferred to real-world applications. 

To summarize, the key contributions of this work are as follows: (1) we introduce the first scalable safe reinforcement learning (safe RL) solution trained in simulation and successfully transferred to a real-world fleet; (2) our approach adapts the Recovery RL framework with a series of enhancements specifically tailored for the self-driving car (SDC) domain.; and (3) we demonstrate results on both a public dataset and a real-world fleet.

The rest of this document is organized as follows: In Section~\ref{sec:related-work} we briefly overview the related work. In Section~\ref{sec:cimrl} we describe our approach and the corresponding architecture in the context of the Recovery RL framework. In Section~\ref{sec:experiments} we describe our experimental setup and present the main results in simulated environment and and evaluate transfer to a real autonomous flee. Finally, in Section~\ref{sec:conclusions} we conclude and present a few future research directions.

\section{Related Work}
\label{sec:related-work}





There has been a vast body of work concerning safety in Reinforcement Learning.
One prominent approach to address safety concerns is modeling the problem using the Constrained Markov Decision Process (CMDP)~\cite{altman2021constrained}.
In this framework, the agent aims to maximize its long-term task objective while satisfying the specified safety constraints.
One technique to solve CMDP is to reshape the task objective by mixing it with the constraint cost~\cite{DBLP:journals/corr/AchiamHTA17} or approximating it within the local region of the current policy (trust region methods)~\cite{DBLP:journals/corr/abs-1805-11074}.
Another approach is to leverage the idea of ``shielding"~\cite{alshiekh2018safe, brunke2022safe, DBLP:journals/corr/abs-2010-15920}, where a shielding layer corrects the action proposed by the agent to remain within the safe state-action space specified by the safety constraints. 
Our method is most relevant to Recovery RL~\cite{DBLP:journals/corr/abs-2010-15920}, which proposes a framework where an agent learns to optimize the task policy while utilizing a (learned) recovery policy that guides the agent back to a safe state when it violates the safety constraints defined in the problem.
This hybrid policy aims to ensure that the agent can operate safely within an environment, effectively mitigating risks and improving the robustness of RL applications in real-world scenarios. In contrast to this work, our approach is not generative in terms of actions; instead, we optimize over a predefined set of motion proposals generated by a previously learned motion generator, such as behavior cloning (BC) policies. These pre-generated motions are generally more likely to remain within safety zones and achieve higher task return while ensuring safer policy exploration in the task space.
Our core RL algorithms are based on the Soft Actor-Critic (SAC)~\cite{DBLP:journals/corr/abs-1812-05905} algorithm. More specifically, since we optimize over a finite set of motion proposals we adopted SAC for discrete action settings~\cite{DBLP:journals/corr/abs-1910-07207}.
The work based on risk measurements is an alternative method~\cite{NIPS2015_024d7f84,greenberg2022efficient} widely used to model risks or safety.

A vast body of work in robotics motion planning has focused on a similar trajectory selection and generation problem, where a library of trajectories is generated offline, selected, and composed during runtime.
By discretizing the large action space into a finite set of trajectory primitives, these methods can achieve reasonable performance while maintaining an acceptable runtime~\cite{fox1997dynamic, pivtoraiko2009differentially, dey2012efficient}.
Recent methods in this line of work augment trajectory primitives with ``funnels"~\cite{tedrake2010lqr, majumdar2017funnel, singh2023robust}, where each funnel is associated with a feedback controller, and the funnel size represents motion uncertainty.
They ensure the safety of the robotic system by selecting the funnels (and the associated controllers) with the smallest cost while accounting for uncertainties in motion~\cite{alterovitz2008stochastic}.
Similar techniques use Hamilton-Jacobi-Bellman (HJB) Reachability analysis \cite{kousik2020bridging, kousik2017safe} to precompute a safe set of trajectory parameters offline and select the best trajectory from the set online.
While funnel-based motion primitives offer analytical safety guarantees concerning the misalignment between the physics model used for planning and the real world, they do not address uncertainties in other agents' decisions.
In theory, these uncertainties could be embedded into funnel-based primitives;
However, identifying uncertainties of other agents is highly challenging, unlike system identification for physical systems.
In contrast, our method trains a closed-loop safety critic using model-free RL by simulating many real-world driving scenarios, bypassing the necessity of explicitly identifying such an uncertainty model.


Our work is also related to learning and planning in hierarchical reinforcement learning~\cite{sutton1999between,NIPS1997_5ca3e9b1,DBLP:journals-corr-cs-LG-9905014l,kaelbling2011hierarchical}
, specifically within the framework of options \cite{sutton1999between}. We can view the task and recovery policies as two separate options where we essentially learn the policies and the termination conditions for each. The high-level policy is a fixed policy that switches between these two options depending on the safety criteria and regions.


\cite{lu2023imitation} proposes an approach (BC-SAC) that addresses the limitations of pure imitation learning (IL) by integrating it with reinforcement learning (RL). The hybrid approach combines the strengths of IL and RL, allowing the system to learn from both expert demonstrations and its own experiences, thereby improving driving behavior realism which emerges from  IL; and robustness and safety using the closed loop feedback control in RL, in diverse driving scenarios. Our approach differs from this work as follows: (1) BC-SAC directly optimizes a joint IL+RL objective by combining the (offline) expert data likelihood augmented by a (online) reward optimization within an RL framework. In contrast, our approach decouples IL and RL, and performs motion planning using the motions generated by a previously learned IL component as the temporally extended actions in a hierarchical RL framework; (2) BC-SAC does not explicitly guarantee multi-criteria safety concerns which are very important in SDC. In contrast, we utilize recovery RL within a safe RL framework to establish safety guarantees in this domain.


\section{Combining IMitiation and Reinforcement Learning (CIMRL)}
\label{sec:rlms-algorithm}
\label{sec:cimrl}
\label{sec:rlms}
This section presents a comprehensive overview of our CIMRL system, including the algorithm, model architecture, and distributed training system.

\subsection{CIMRL Model}
\label{sec:cimrl}
The CIMRL system adapts the Recovery-RL framework~\cite{DBLP:journals/corr/abs-2010-15920}  which prioritizes policy safety while maintaining steady progress along the ego vehicle’s route. More specifically CIMRL is a constrained MDP (CMPD)~\cite{DBLP:journals/corr/AchiamHTA17} at the core which can be defined as a tuple $\mathcal{M} \equiv \langle \mathcal{S}, \mathcal{A}, \mathcal{T}, \mathcal{R}, \gamma_{task}, \mathcal{C}, \gamma_{risk} \rangle$ where $\mathcal{S}$ is a set of states, $\mathcal{A}$ is the set of actions, $\mathcal{T}(\mathcal{S}, \mathcal{A}, \mathcal{S}') \mapsto [0, 1]$ is the transition dynamics, $\mathcal{R}$ and $\gamma_{task}$ are task reward and the corresponding discounting factor, $\mathcal{C}$ and $\gamma_{risk}$ are the risk cost and the corresponding discounting factor\footnote{Without loss of generality, we assume a single risk model, though this can be easily extended to accommodate multiple risks.}. Given a policy $\pi$ the task and risk returns are defined as {\small $\mathcal{J}_{task}(\pi) = \mathbb{E}_{\tau \sim \pi} \left[ \sum_{t=0}^{\infty} \gamma^{t}_{task} R(s_t, a_t) \right]$} and {\small $\mathcal{J}_{risk}(\pi) = \mathbb{E}_{\tau \sim \pi} \left[ \sum_{t=0}^{\infty} \gamma^{t}_{risk} C(s_t, a_t) \right]$} where {\small $\tau \equiv \{s_0, a_0, s_1, \hdots\}$} denotes a sample trajectory generated by the policy $\pi$. Similarly, state-action value functions associated with the task and risk for a policy $\pi$ can be defined as {\small $Q^{\pi}_{task}(s, a) = \mathbb{E}_{\tau \sim \pi} \left[ \sum_{t=0}^{\infty} \gamma^{t}_{task} R(s_t, a_t) | s_0=s, a_0=a\right]$} and {\small $Q^{\pi}_{risk}(s, a) = \mathbb{E}_{\tau \sim \pi} \left[ \sum_{t=0}^{\infty} \gamma^{t}_{risk} C(s_t, a_t) | s_0=s, a_0=a)\right]$}.\\


The optimization objective of CMDP can be defined as:
{\small
\begin{align}
    \pi^* &\equiv \arg\max_{\pi} \mathcal{J}_{task}(\pi) \quad \text{s.t.} \quad  \mathcal{J}_{risk}(\pi) < \epsilon_{risk} \nonumber\\
    &\text{or similarly:} \nonumber\\
    \pi^* &\equiv \arg\max_{\pi} Q^{\pi}_{task}(s, a) \quad \text{s.t.} \quad  \mathbb{E}_{(s, a)} \left[Q^{\pi}_{risk}(s, a)\right] < \epsilon_{risk}
    \label{eq:cmpd-opt}
\end{align}
}

Optimizing the objective in Equation~\ref{eq:cmpd-opt} presents a set of challenges in domains like self-driving cars (SDC), which involve multi-objective optimization with conflicting goals, particularly when a single flat policy structure is used. These objectives include maximizing task rewards, such as progress along the ego route, while minimizing risks, such as avoiding collisions. Recovery RL~\cite{DBLP:journals/corr/abs-2010-15920} is a specialized CMDP algorithm designed to approximate a solution to the optimization problem outlined above by utilizing a hierarchical policy optimization. First, we define agent {\em safe zone} as\footnote{To simplify notation, we omit the dependency of $\Omega(\mathcal{S})$ on the risk value network $Q^{\pi}_{risk}$ and assume it will be understood from the context.}:
\begin{equation}
    \Omega(\mathcal{S}) := \{ s \in \mathcal{S} \, | \,  \exists a \in \mathcal{A}(s) \; \text{s.t.} \; Q^{\pi}_{risk}(s, a) \leq \epsilon_{risk} \}
\end{equation}
This zone essentially defines a set of states where at least one action exists that, when taken, does not lead to an unsafe state (i.e., does not violate the risk threshold). 
We then define the top level policy $\pi_{comb}$ (short for combined policy) as\footnote{For clarity we use the more intuitive notation $\pi_{recov}(a | s)$ instead of $\pi_{risk}(a | s)$.}:
\begin{align}
\pi_\text{task}(a | s) &= \arg\max_{\pi} Q^{\pi}_\text{task}(s, a), \; \forall s \in \Omega(\mathcal{S})\nonumber\\
\pi_{recov}(a | s) &= \arg\min_{\pi} Q^{\pi}_{risk}(s, a)\nonumber\\
\pi_{comb}(a | s) &=
\begin{cases}
    \pi_\text{task}(a | s), & \text{if } s \in \Omega(\mathcal{S})\\
    \pi_{recov}(a | s), & \text{else}
\end{cases} \nonumber\\
\end{align}
which utilizes two concurrently trained policies: $\pi_\text{task}$ and $\pi_{recov}$ (short for {\em recovery}). The task policy, $\pi_\text{task}$, is responsible for maximizing the task critic, while the recovery policy, $\pi_{recov}$, focuses on minimizing a separate risk critic, which estimates potential constraint violations. This structure enables effective management of conflicting objectives. CIMRL further extends Recovery RL by introducing the following enhancements:

\subsubsection{Actions}
\label{sec:cimrl-action}
Most prior work explores RL with a fixed action space (e.g. a globally consistent $\leftarrow \uparrow \downarrow \rightarrow
$ action space in Atari games) or in continuous action spaces~\cite{DBLP:journals/corr/abs-1812-05905}. In CIMRL, we consider the case of a state-dependent action space where the possible actions at every state are dynamically produced by a
an auxiliary motion prior generator component as depicted in Figure~\ref{fig:cimrl-arch}. More formally, the action space $\mathcal{A}(s)$ for a given state $s$ is defined by the Motion Generator $\text{MG}: \mathcal{S} \mapsto \mathcal{T} \equiv \{\tau_1, \tau_2, \hdots\}$ where $\tau_i$ is a motion trajectory (e.g., $\tau_i = \{(x^i_t, y^i_t, \text{yaw}^i_t)\}^{k-1}_{t=0}$ for some fixed horizon $k$). re 1). The choice of the
motion prior generator is left to the user and is not specific to our approach. In general, this module could consist of a combination of a pretrained ML trajectory generator (e.g., \cite{shi2022motion}) and/or a classical planner trajectory generator (e.g., \cite{LaValle1998RapidlyexploringRT}). Our method is entirely agnostic to the trajectory generation process and remains equivariant to the order of motion priors presented to the learner architecture and additionally supports variable numbers of motion plans at a given state. 

\begin{table*}[h]
\label{exp:waymax-exp}
\begin{center}
\begin{tabular}{|c | c | c|c|} 
 \hline
\textbf{Method} & \textbf{Log ADE (m) $\downarrow$ } & \textbf{Collision Violation (\%) $\downarrow$ } & \textbf{Offroad Violation (\%) $\downarrow$ } \\
 \hline
 MTR (highest prob) & $29.93 $& 41.29 & 14.22 \\
MTR (sample) & 30.78 & 43.28 & 15.39 \\
SAC & 0.58 & 18.03 & 24.73 \\
CIMRL (Ours) & 1.62 & 16.76 & 21.65 \\
\hline
\end{tabular}
\end{center}
\label{table:comparison-waymax}
\caption{Performance Comparison of Different Methods (Waymax)}
\end{table*}
\subsubsection{Suppressed Task Value}
\label{sec:suppressed-task-value}
Oftentimes, constrained safe RL algorithms can result in sharp discontinuities in the policy, particularly near the boundaries of constraint violations. To minimize the potential for abrupt changes in the policy near safety-critical states, we adopt a strategy similar to \cite{zhou2024multiconstraint} to adaptively suppress the task $Q$ value using the risk $Q$ value as follows:
\begin{equation*}
\small
Q_{\text{task}\downarrow}(s, a) = \frac{Q_\text{task}(s, a)} {f(Q_\text{risk}(s, a))}
\end{equation*}
where $f(.)$ is is a monotonically increasing partition function of $Q_\text{risk}(s, a)$ (e.g., an exponential function such as $\exp(.)$)\footnote{Note that we use the new notation $Q_{\text{task}\downarrow}(s, a)$ (with the down arrow) to denote the suppressed task value function.}. In CIMRL, The task policy aims to maximize this suppressed task value $Q_{\text{task}\downarrow}(s, a)$. This facilitates a smooth interpolation between the conflicting objectives of maximizing task reward while minimizing risk severity. It also modulates the sensitivity to the applied risk threshold $\epsilon_{\text{risk}}$ in the combined policy as described in Section~\ref{sec:cimrl}, especially in the nearby vicinity of safety-critical states. In practice, we have found it beneficial to suppress $Q_\text{task}(s, a)$ only when $Q_\text{risk}(s, a)$  
exceeds a certain threshold. For example, we can utilize a quantized exponential function $f(Q_{\text{risk}}(s, a)) = \exp(\tau \lfloor \frac{Q_{\text{risk}}(s, a)}{\rho} \rfloor$)
for some quantization parameter $\rho$ and temperature $\tau$ which makes the suppressed objective less sensitive to the inherent noise in the $Q_\text{risk}(s, a)$ calculation.

\subsubsection{D-SAC with Tree Backup}
\label{sec:d-sac}
In self-driving tasks, the feedback is typically sparse, especially for risk events such as collisions.
The one-step backup target estimation used in the original Recovery RL may struggle to propagate the delayed risk feedback accurately, resulting in slow and biased learning for the agent.
To overcome this limitation, in CIMRL, we combine D-SAC with the Tree-Backup algorithm \cite{Asis2017MultistepRL}, an off-policy $N$-step method, to estimate the value function targets.
The Tree Backup algorithm starts from the final step of the episode and propagates the target values backward in time.
Since most self-driving simulators limit the simulation to a fixed number of steps~\cite{gulino2023waymax}, which may not result in a driving scene termination, we use bootstrapped estimation rather than the final reward for computing the last step target values
\begin{equation}\label{eqn:last-step-target}
    G^T_\text{task} = Q_{\text{task}\downarrow}(s_T, a_T), \space
    G^T_\text{risk} = Q_\text{risk}(s_T, a_T),
\end{equation}
where $G^T_\text{task}$ and $G^T_\text{risk}$ denote the target values for task and risk state-action value functions at time step $T$, respectively.
The task target value at $T-1$ is given by one-step backup using the Bellman equation
\begin{equation}\label{eqn:second-to-last-step-task-target}
G^{T-1}_\text{task} = r_\text{task}(s_{T-1}, a_{T-1}) + \gamma_\text{task}V_\text{task}(s_{T}),
\end{equation}
where $\gamma_\text{task}$ is the discount factor for the task value function and $G_\text{task}$ denotes the task target value.
The next state value is given by $V_\text{task}(s_{t+1})=\sum_{a}\pi_\text{task}(a|s_{t+1})Q_\text{task$\downarrow$}(s_{t+1}, a)$, and similar for the risk value $V_\text{risk}(s_{t+1})$.
The risk target value is computed by
\begin{equation}\label{eqn:second-to-last-step-risk-target}
\small
G^{T-1}_\text{risk} =
\begin{cases}
    C(s_{T-1},a_{T-1}), & \text{if } Q^{\pi}_\text{risk}(s_{T-1}, a_{T-1}) \leq \epsilon_\text{risk} \\[6pt]
    \gamma_\text{risk}V_\text{risk}(s_T), & \text{otherwise}
\end{cases}
\end{equation}
where $V_\text{risk}(s_T)$ is computed similarly as $V_\text{task}(s_T)$.
When computing risk targets, we disregarded future risk severity beyond the initial failure event determined by the risk value function, aligning with the approach in Recovery RL.
For timesteps $t < T-1$, the target values are computed recursively based on the previous ones.
The target for $Q_\text{task}$ is
\begin{equation}\label{eqn:task-target}
\small
G^{t}_\text{task} = r_\text{task}(s_t, a_t) + \gamma_\text{task}\left(\pi_\text{task}(a_{t+1}|s_{t+1})G_\text{task}^{t+1} + \hat{Q}_\text{task$\downarrow$}\right),
\end{equation}
where $\hat{Q}_\text{task$\downarrow$} = \sum_{a \neq a_{t+1}}\pi_\text{task}(a | s_{t+1})Q_\text{task$\downarrow$}(s_{t+1}, a)$ is the off-policy target, estimated as the expected Q-value of off-policy actions weighted by the corresponding action probabilities.
The risk target can be computed similarly 
\begin{equation}\label{eqn:risk-target}
\small
\begin{aligned}
G^{t}_\text{risk} &= 
\begin{cases}
    C(s_{t},a_{t}), & \text{if } Q^{\pi}_\text{risk}(s_{t}, a_{t}) \leq \epsilon_\text{risk}\\[6pt]
    \begin{array}{@{}l@{}}
    \gamma_\text{risk}\left(\pi_\text{risk}(a_{t+1} | s_{t+1})G_\text{risk}^{t+1} + \hat{Q}_\text{risk}\right), 
    \end{array}
    & \text{otherwise}
\end{cases}
\end{aligned} 
\end{equation}
where $\hat{Q}_\text{risk} = \gamma_\text{risk}\sum_{a \neq a_{t+1}}\pi(a | s_{t+1})Q_\text{risk}(s_{t+1}, a)$ is the off-policy estimate for the risk target.
The target values computed through Eq.~\eqref{eqn:last-step-target} to Eq.~\eqref{eqn:risk-target} are used as the regression targets for both the task and risk networks.
The loss for the value functions is
\begin{equation}
\begin{aligned}
\mathcal{L}^{\mathcal{Q}}_\text{task} &= \mathbb{E}_{(s_{t}, a_{t}) \sim \mathcal{D}_\text{task}}\left[\left(Q_\text{task}(s_t, a_t) - G_\text{task}^t\right)^2\right] \\ 
\mathcal{L}^{\mathcal{Q}}_\text{risk} &= \mathbb{E}_{(s_{t}, a_{t}) \sim \mathcal{D}_\text{recov}}\left[\left(Q_\text{risk}(s_t, a_t) - G_\text{risk}^t\right)^2\right],
\end{aligned}
\end{equation}
where $\mathcal{D}_\text{task}, \mathcal{D}_\text{recov}$ are the replay buffers storing the experiences for the task and recovery policies, respectively.

Since we operate in a discrete setting, we can update the policies by directly minimizing the Kullback-Leibler (KL) divergence between the policy distribution and a Boltzmann (softmax) distribution induced by the discrete Q-functions
\begin{equation}\label{eq:policy-optimization}
\begin{aligned}
\mathcal{L}^{\mathcal{\pi}}_\text{task} &= D_\text{KL} \left[\pi_\text{task}(a|s) \parallel \mathrm{softmax}(Q_\text{task}^\pi(s, a)) \right]\\
\mathcal{L}^{\mathcal{\pi}}_\text{recov} &= D_\text{KL} \left[\pi_\text{recov}(a|s) \parallel \mathrm{softmax}(-Q_\text{risk}(s, a)) \right].
\end{aligned}
\end{equation}
It’s important to note that the recovery policy is trained to minimize risk achieved by taking the negative of the risk utility. The pseudo-code for the CIMRL algorithm is presented in Algorithm~\ref{alg:cimrl}.

\begin{algorithm}
\caption{CIMRL Algorithm}\label{alg:cimrl}
\begin{algorithmic}[1]
\State {\bf Inputs}: Motion Generator $\text{MG}: \mathcal(S) \mapsto \mathcal{T} \equiv \{\tau_1, \tau_2, \hdots\}$, Reward function: $\mathcal{R}$, Risk function: $\mathcal{C}$, Risk Threchold: $\epsilon_{risk}$
\State $\mathcal{D}_{task} \gets \emptyset, \mathcal{D}_{recov} \gets \emptyset$;
\While{training}
\For{$i$ in $\{1, \hdots, N\}$}\Comment{$N$ rollouts}
\State  $s \sim P(s_0)$
\For{$j$ in $\{1, \hdots, H\}$}\Comment{rollout of horizon $H$}
    \State $\mathcal{A}(s) \gets \text{MG}(s)$ \Comment{Generate motion priors in state $s$}
    \If{$s \in \Omega(\mathcal{S})$} \Comment{If $s$ is safe, optimize task}
        \State $a \gets \pi_{\text{task}}(a | s), \; a \in \mathcal{A}(s)$ \Comment{}
        \State Observe the reward $r$, and next state $s'$
    \State $\mathcal{D}_{task} \gets \mathcal{D}_{task} \cup (s, a, r, s')$
    \Else \Comment{If $s$ is unsafe, minimize risk}
        \State $a \gets \pi_{\text{recov}}(a | s), \; a \in \mathcal{A}(s)$ \Comment{}
        \State Observe the risk $c$, and next state $s'$
        \State $\mathcal{D}_{recov} \gets \mathcal{D}_{recov} \cup (s, a, c, s')$
    \EndIf
    \State $s \gets s'$
\EndFor
\EndFor
\State Train $\pi_{task}$ on $\{\mathcal{L}^{\mathcal{Q}}_\text{task}, \mathcal{L}^{\pi}_\text{task} \; | \; \mathcal{D}_{task}\}$
\State Train $\pi_{recov}$ on $\{\mathcal{L}^{\mathcal{Q}}_\text{recov}, \mathcal{L}^{\pi}_\text{recov} \; | \; \mathcal{D}_{recov}\}$
\EndWhile
\end{algorithmic}
\end{algorithm}

\subsection{Model Architecture}
\label{sec:model-architectures}
The CIMRL model, as illustrated in Figure \ref{fig:cimrl-arch}, is based on an actor-critic architecture where there is a shared state encoder backbone with separate heads to predict the task and risk critic, as well as the task and recovery policy.

\subsubsection{Encoder}
\label{sec:encoder}
States and actions are encoded separately using neural networks, then concatenated before being passed to the learning algorithm (Figure~\ref{fig:cimrl-arch}). The choice of the state encoder is left to the user
and is not specific to our framework. However, in general, a commonly used Transformer architecture \cite{DBLP:journals/corr/VaswaniSPUJGKP17} could be leveraged to encode track history and map for state embedding at each timestamp. 
For actions, the model takes as input a variably sized set of N-second motion plans and projects them into an embedding space using MLPs or transformers which is then concatenated with the aforementioned state embedding.


\subsubsection{Decoder}
\label{sec:decoder}
Using the state and action embeddings, we concatenate them to predict multiple outputs, as detailed below. For training each of the decoder networks, we utilize episodic sequences (e.g., 30-seconds of real world driving).

\par {\em Task and Risk Critic Networks:} We build two critic networks, a task critic $Q_\text{task}$ and a risk critic $Q_\text{risk}$. Both of these networks are implemented with MLPs on top of the concatenated state and action embeddings.
Each motion plan is associated with corresponding task and risk values estimated by these critics.

\par {\em Task and Recovery Policy Networks:} We build two policies networks, $\pi_\text{risk}$ and $\pi_\text{recov}$. Each of these networks is also modeled with MLPs. We output a policy probability for each valid action for each of the policy networks.

\subsubsection{Model Training}
\label{sec:training-details}
Training a reinforcement learning (RL) models is challenging because both the task and risk targets are typically derived from bootstrapped Q-value predictions. Without a pretrained model from imitation learning (IL) or offline RL algorithms~\cite{DBLP:journals/corr/abs-2110-06169}, the initial policy and critic estimates are often random, which can lead to inaccurate predictions and potentially destabilize the training process. In practical applications, both task and risk targets carry significant real-world semantics. For instance, knowing the speed limit for the autonomous driving (AD) system allows us to estimate the maximum progress reward per simulation step, as well as establish prediction bounds for the progress task target using $R_{\text{max}} = \frac{\delta^*}{1 - \gamma}$ where $\delta^*$ denotes the maximum progress and $\gamma$ is a discounting factor. As for risk, if we are minimizing risk severity, we can safely assume and compute the risk severity as a non-negative value, and we can apply a softplus function $f(x) = \log(1 + \exp(x))$ to the raw network’s output. In practice, we found that this careful initialization for the task and risk Q value prediction layer can enable us to train the RL model from scratch and have a stable training process.


\section{Experiments}
\label{sec:experiments}
The goal of the experimental evaluation is to demonstrate that the successful application of RL on discrete action spaces can outperform baselines relying on imitation and continuous action spaces alone. We compare our method to baselines based on common state-of-the-art methods for the driving domain and provide comparisons on both open-source and proprietary environments to demonstrate the generalizability of our method.

\subsection{Waymax}
\subsubsection{Setup}
The Waymax simulator \cite{gulino2023waymax} is a differentiable driving simulator built on top of the standard Waymo Open Motion Dataset and covers a wide range of scenarios ranging in difficulty taken from real-world data. Waymax provides several distinct metrics for evaluation namely: Average Displacement Error (ADE) from the logged ego behavior, Offroad Rate which measures the tendency for the ego to drive off the context map, and Collision Rate. We use an action repeat of $5$ in order to both train and evaluate each method which results in a control frequency of $2$hz. We use MTR~\cite{shi2023mtr} as the motion trajectory generator. Since the trajectories output by the MTR model are non-smooth, they produce a noisy yaw signal. We smooth the noise in yaw by applying a low pass filter on the computed yaw before executing the action. In Table \ref{table:comparison-waymax} we compare the CIMRL with the naive application of MTR as a trajectory generator. Our method strongly outperforms the cloning baseline represented by MTR across ADE as well as collision violations while remaining competitive in offroad violations. 

\subsubsection{Result}
The results are shown in Table~\ref{exp:waymax-exp}.
In general, CIMRL can augment and significantly improve upon the closed-loop behavior of an open-loop policy resulting in much lower divergence and collisions. However, the rate of violations remains high and can be explained by the specific architecture of the MTR model. In closed loop, it is common to accumulate drift over time, and in the case of MTR, the initial goal waypoints are rotated into the ego frame. Error in yaw over time will then lead to offroad goals due to this rotation making it challenging for any policy to recover. This result highlights the differences between open and closed loop benchmarks and suggests that open loop performance might not correlate well with closed loop behavior. 

\subsection{Real World}
We train our method on hundreds of thousands of scenes from both real onroad data as well as synthetic leveraging trajectories from both a learned model as well as heuristically generated backup plans. Our evaluation dataset consists of $30$k scenes covering a wide range of ODD as well as interaction types. These scenes are mined in order to contain challenging scenes for both open loop prediction models as well as capturing long-tail interactions. In Table \ref{table:comparison-inhouse} we compare the naive behavior Cloning (BC) implementation with the CIMRL on top of either only BC trajectories or BC trajectories combined with the heuristic planner proposals.

\begin{table}[h]
\begin{center}
\begin{tabular}{|c c c|} 
 \hline
\textbf{Method} & \textbf{Collision Violation (\%) $\downarrow$ } & \textbf{Stuck (\%) $\downarrow$} \\
\hline
BC & 100.00 & 100.00 \\
CIMRL (BC) & 82.22 & 54.54 \\
CIMRL (BC+Heuristics) & 87.64 & 16.98 \\
\hline
\end{tabular}
\end{center}
\label{table:comparison-inhouse}
\caption{Performance Comparison of Different Methods in real world (In-house)}
\end{table}

On proprietary datasets, it is clear that CIMRL can improve upon a purely cloned baseline policy on safety (collision) as well as task completion rate (decreased stuck). The addition of heuristic based plans expands to action space making exploration more difficult, but also enables the injection of recall that might be missing from the learned models. Additionally, we train and evaluate without reactive agent behaviors so policies are not expected to reach 0 collision rate since any small deviation from what happened when the data was collected can lead to collisions later in the scene despite optimal behavior.   

With this evaluation in simulation, we are able to confidently run policies trained with CIMRL on public roads.

\section{Conclusions and Future Directions}
\label{sec:conclusions}
In this work, we presented a safe reinforcement learning (RL) framework for motion planning in self-driving cars which can be built on top of an imitation learning trajectory generator --- and generally speaking, can use any trajectory generator source (e.g., heuristic planners, geometric-based samplers, etc.). Our approach reduces the complexity of the action space to a narrow subset of top motions generated by state-of-the-art motion generators and employs reinforcement learning to select a long-horizon optimization objective. Furthermore, we proposed embedding this framework within a safe RL super-set to incorporate explicit risk severity modeling, which is crucial in any self-driving car algorithm. One interesting future direction would be to use the closed loop feedback and further refine the fixed motions generated by the motion generator component.

\paragraph*{Acknowledgements:} We would like to thank the many people who contributed to this work during design, development, and writing. Specifically we would like to thank Zixun Zhang, Jinrui Huang, and Ethan Tang for their support in developing the infrastructure to support development. Jin Ge for her insight. Kai Ang for their early contributions.

\bibliography{references}{}
\bibliographystyle{plain}

\end{document}